# Copycat CNN: Stealing Knowledge by Persuading Confession with Random Non-Labeled Data

Jacson Rodrigues Correia-Silva*, Rodrigo F. Berriel, Claudine Badue, Alberto F. de Souza, Thiago Oliveira-Santos
Universidade Federal do Espírito Santo, Brazil
*Email: jacson.silva@ufes.br

*Abstract*—In the past few years, Convolutional Neural Networks (CNNs) have been achieving state-of-the-art performance on a variety of problems. Many companies employ resources and money to generate these models and provide them as an API, therefore it is in their best interest to protect them, i.e., to avoid that someone else copy them. Recent studies revealed that state-of-the-art CNNs are vulnerable to adversarial examples attacks, and this weakness indicates that CNNs do not need to operate in the problem domain (PD). Therefore, we hypothesize that they also do not need to be trained with examples of the PD in order to operate in it.

Given these facts, in this paper, we investigate if a target black-box CNN can be copied by persuading it to confess its knowledge through random non-labeled data. The copy is two-fold: *i)* the target network is queried with random data and its predictions are used to create a fake dataset with the knowledge of the network; and *ii)* a copycat network is trained with the fake dataset and should be able to achieve similar performance as the target network.

This hypothesis was evaluated locally in three problems (facial expression, object, and crosswalk classification) and against a cloud-based API. In the copy attacks, images from both non-problem domain and PD were used. All copycat networks achieved at least 93.7% of the performance of the original models with non-problem domain data, and at least 98.6% using additional data from the PD. Additionally, the copycat CNN successfully copied at least 97.3% of the performance of the Microsoft Azure Emotion API. Our results show that it is possible to create a copycat CNN by simply querying a target network as black-box with random non-labeled data.

## I. INTRODUCTION

Convolutional Neural Network (CNN) is a category of Deep Neural Network (DNN) that has recently been achieving state-of-the-art performance on a variety of problems, such as visual classification and recognition, object detection, and others. Given its outstanding performance in a wide range of tasks, many companies started to offer solutions based on these neural networks. In fact, several companies already have APIs (Application Programming Interfaces) that give access to their deep learning models. Most of them provide (e.g., Amazon AWS, Microsoft Azure, Google Cloud, BigML, etc.)

Scholarships of Productividade on Research (grants 311120/2016-4 and 311504/2017-5) supported by Conselho Nacional de Desenvolvimento Científico e Tecnológico (CNPq, Brazil); and a scholarship supported by Coordenação de Aperfeiçoamento de Pessoal de Nível Superior (CAPES, Brazil).

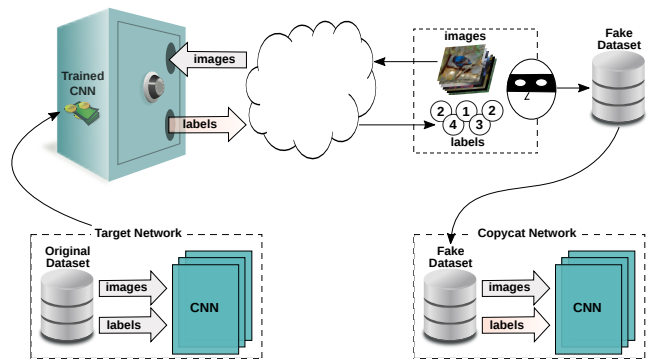

Fig. 1: On the left, the target network is trained with an original (confidential) dataset and is served publicly as an API, receiving images as input and providing class labels as output. On the right, it is presented the process to get stolen labels and to create a fake dataset: random natural images are sent to the API and the labels are obtained. After that, the copycat network is trained using this fake dataset.

cloud-based services to customers allowing them to offer their own models as an API. Because of the resources and money invested in creating these models, it is in the best interest of these companies to protect them, i.e., to avoid that someone else copy them.

Some works have already investigated the possibility of copying models by querying them as a black-box. In [1], for example, the authors showed how to perform model extraction attacks to copy an equivalent or near-equivalent machine learning model (decision tree, logistic regression, SVM, and multilayer perceptron), i.e., one that achieves close to 100% agreement on an input space of interest. In [2], the authors evaluated the process of copying a Naive Bayes and SVM classifier in the context of text classification. Both works focused on general classifiers and not on deep neural networks that require large amounts of data to be trained leaving the question of whether deep models can be easily copied. Although the second uses deep learning to steal the classifiers, it does not try to use DNNs to steal from deep models. Additionally, these works focus on copying by querying with problem domain data.

In recent years, researchers have been exploring some intriguing properties of deep neural networks [3], [4]. More specifically, investigating why someone can cause the network to misclassify an image by applying a certain imperceptible perturbation to the image. These properties makes state-of-the-art CNNs vulnerable to adversarial examples attacks [5]–[8]. In this type of attack, an image of a certain class, and predicted as such by a trained CNN, is modified by adding some non-visible noise so that the network prediction is changed to a different label and the attacker gets the system to perform in a directed manner. This property indicates that CNNs do not need to operate the problem domain. Therefore, we hypothesize that they also do not need to be trained with examples of the problem domain in order to operate in the problem domain. In the counter side, works have also been done to try to avoid such attacks [9], [10], but, in general, state-of-the-art CNNs are still vulnerable.

Given the imperfection of these models and the enormous amount of data required to train such models, questions arise on whether they can be easily copied when made available as a black-box.

In this work, we present a novel and simple, yet powerful, method to copy a black-box CNN model by persuading it to confess its knowledge through random non-labeled data. The copying is performed in two steps (Figure 1). First, the target network is queried with random non-labeled data and the pairs of input image and output label are used to create a fake dataset with the knowledge of the network. Second, a copycat network is trained with the fake dataset and should be able to achieve similar performance as the target network.

This method was evaluated in three problems (facial expression, object, and crosswalk classification). In addition, a copy attack was performed against a cloud-based API from Microsoft. The copy attacks were performed with two different types of random data (non-problem domain and problem domain). These two types of data enables answering whether a CNN requires problem specific training data in order to operate properly in a specific domain. Our results showed that all copycat networks were able to achieve at least $93.7\%$ of the performance of the original models. When the training was performed with additional images from the problem domain, the copycat networks were able to achieve at least $98.6\%$ of the performance of the original models. An experiment with the Microsoft Azure Emotion API also showed that the proposed method can successfully ($+97\%$) copy the performance of a black-box model. In conclusion, the study demonstrated a weakness of CNN models that can cost a lot for companies, i.e., showed that it is possible to create a copycat CNN by simply querying a target network as black-box with random non-labeled data.

To the best of our knowledge, this is the first time the process of copying a deep CNN (black-box) model using random unlabeled data is investigated. Our findings open up doors for future investigations in the field of transfer learning. Our results showed that the copy is possible even in real world problems since the method was evaluated using different combinations of databases and problem domains to copy the performance of a target network.

The main contributions of this paper can be listed as follows:
- A technique to copy black-box CNN models;
- The finding that a model can be copied using ordinary natural and non problem related images;
- Extensive evaluation shows it is viable on different problems and big datasets.

## II. RELATED WORKS

There are few works in the literature when it comes to the investigation of attacks to black-box models, even less involving deep models. Some of these works are presented in this section.

In [7], the authors used a black-box target DNN to provide labels to an augmented version of the MNIST database. After that, a substitute DNN was generated, using this dataset, achieving over $80\%$ of accuracy. The substitute DNN is, then, used to craft adversarial samples against the target DNN. The intention was approximate the decision boundaries of the target DNN just to craft adversarial examples. A refinement to this technique was proposed in [5]. There, the authors transfered the knowledge from machine learning classifiers (DNN, logistic regression, SVM, decision tree, nearest neighbor, and ensembles) to a deep model (again, used to mimic the decision boundaries of the original classifier). They also used MNIST dataset. In addition, the substitutes were able to approximate only about $77\%$ to $83\%$ of accuracy in the test set, with $48\%$ for the decision trees.

Another related paper is [2], where a deep learning classifier is trained with input data labeled by a target classifier. The authors generated two classifiers for text classification based on a binary dataset: Naive Bayes and SVM. To steal these classifiers, the authors generated two deep learning classifiers that were able to produce high fidelity labels of $97.9\%$ and $97.44\%$ of accuracy, respectively, similar to the target models.

In [1], the authors use partial knowledge of models and equation solving to train similar models with the goal of achieving an equivalent or near-equivalent machine learning model. Their black-box attack had successful results on decision tree, logistic regression, SVM, and multilayer perceptron against BigML and Amazon Machine Learning. However, given the cost to copy shallow models, their approach seems not to be scalable to DNN models.

Different from these previous works, we investigate a copy attack from a deep model (seen as the target network) to another deep model (which we call copycat network). In addition, our investigation uses real-world datasets from three different problems. Moreover, our method leverage not only data from the same problem domain, but also random natural images found in public datasets and in the Internet. Furthermore, this study works with high-dimensional inputs and deeper neural networks. Lastly, an attack against a public API is also performed.

## III. COPYCAT CONVOLUTIONAL NEURAL NETWORK

The proposed method aims at copying a target network into a copycat version by only performing queries with random natural images. It mainly comprises two main steps: fake dataset generation and copycat network training. In the first step, a fake dataset is generated using random natural images labeled by a given model (target network used as black-box). Then, a copycat network is trained with this fake dataset aiming to copy the performance of the target network.

## A. Fake Dataset Generation

In order to train a copycat network, a dataset is required. We propose to use a fake dataset annotated by the target network. The fake dataset name was chosen because the dataset actually comprises images that are not related to the problem domain or images that are related to the problem domain, but might not be properly labeled. Therefore, the fake dataset is totally different from the one originally used to train the target network.

Initially, suppose that there is a target network (in this case, a CNN) that can be used only as a black-box, i.e., it receives images as input and provides class labels as output. The assumption that the classification space is larger than the one occupied by training examples [3], [4], [11] leads to the assumption that the CNN can and will make mistakes. Nonetheless, to ensure all nuances of the model are captured, a large-scale and varied dataset seems to be required. This dataset can be either made of images from the same problem domain as the target network or made of random natural images. In this work, both approaches as explored. In the first case, it is assumed that the attacker has images from the same problem-domain (PD) of the one used to train the target network. In the second case, it is assumed that the attacker has nothing but access to publicly available large-scale datasets, i.e., random natural images or non-problem-domain (NPD) images. In both cases, original labels (if they are available, which is not required) are assumed to be irrelevant. Therefore, they are discarded.

To generate the fake dataset, the attacker uses the target network itself to automatically label these datasets (PD and/or NPD). These labels generated by the target network are referred in this work as stolen labels (SL). The labeled fake dataset is expected to capture the nuances of the feature space that would enable another network to be trained and achieve a performance close to the target network. As explained before, this assumption is based on the fact that the input image of a CNN can be perturbed by imperceivable noise in order to get the network answer in a certain direction.

Considering the fact that in real applications it is not easy to acquire images of a certain domain, the PD dataset is expected to be smaller than NPD. The non problem domain in the other hand can be freely found on the Internet. Then, when dealing with small databases sizes (e.g., the PD datasets) data augmentation procedures (explained in the Section IV-E) can help increasing the database size in order to better exploit the high-dimensionality space of the target network.

## B. Copycat Network Training

After generating a fake dataset, the training of a copycat network can be performed. In a first step, the attacker would choose a copycat model architecture. Note that the attacker might not know the model architecture of the target network to perform the copy, but this should not hinder the copy of the knowledge to a different model.

In order to assist the comparison process, in this work, the model was fixed to a well known architecture (VGG-16 architecture [12], referred to as VGG only) that, in the performed experiments, was the same for the target and copycat networks.

The VGG was originally created to work with face recognition and, therefore, has a predefined output layer for that specific problem domain. This might also be the case for the architecture chosen, by an attacker, to be the copycat of a target network. Therefore, the output of the chosen model has to be adapted to the problem domain of the target network, i.e., the number of outputs of the copycat has to be changed to match the number of classes handled by the target network.

The literature has shown that, in general, it is a good practice to start training from pre-trained models [13]. Therefore, whenever available, the chosen model should load pre-trained weights to the unchanged layers even if the attacker has a big fake dataset. Since the output was changed to match the problem domain, it has to be initialized with random weights (e.g., using [14]). There are several pre-trained models available nowadays. If available, preference should be given to pre-trained models closer to the problem domain. Otherwise, one should use generic pre-trained models, such as the models pre-trained on the ImageNet.

Finally, the process of training a copycat network consists of finetuning the pre-trained with the fake dataset.

## IV. EXPERIMENTAL METHODOLOGY

This section presents the experimental methodology of creating a copycat from a target black-box CNN. Figure 2 presents an overview of the experimental methodology. The proposed method is compared to two baselines (the original network and an alternative version trained with less data from the problem domain) using a test dataset, while considering three types of copycats (one trained with data not related to the problem domain, one trained with data from the problem domain, and one trained with the former and fine-tuned with the latter). Each of the three copycat networks are evaluated in three different real-world problem types: Facial Expression Recognition (FER), General Object Classification (GOC) and Satellite Crosswalk Classification (SCC).

## A. Test Dataset and Baselines Methods

For all experiments, there is one dataset used for testing the performance (test dataset), and there are two methods used as baselines: *i)* the original target network trained with the original dataset (OD) and *ii)* an alternative version trained with less data of the problem domain with their original labels (PD-OL). The test dataset and baselines are detailed below.

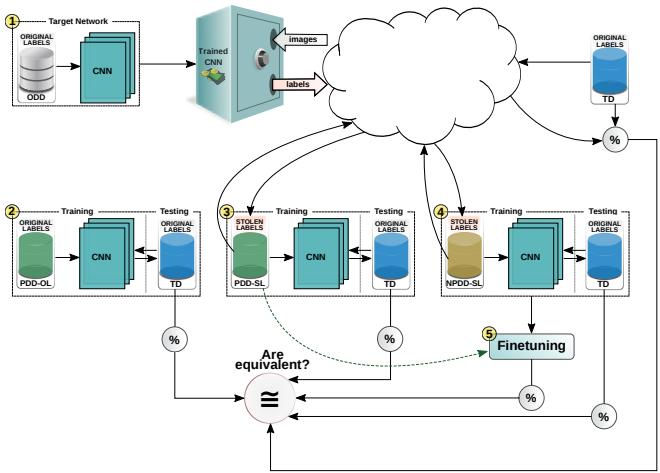

Fig. 2: Experimental methodology to evaluate the copying process. (1) represents the target network trained with original domain dataset, ODD (used as baseline). (2) represents the baseline network trained with a small set of images from the problem domain with original labels, PD-OL. (3) represents the network trained with images from the problem domain with stolen labels, PD-SL. (4) represents the images from the non-problem domain with stolen labels, NPD-SL. (5) represents the finetuning of NPD-SL network with PD-SL dataset, NPD+PD-SL. The accuracy for all networks is obtained on test dataset (TD) and compared with the target network accuracy and with the baseline network accuracy PD-OL.

*1) Test Dataset (TD):* This dataset is used to evaluate the performance of all networks (baselines and copycats). This dataset comprises images from the same problem domain as the target network, but with no overlap. In our experiments, one dataset is defined as the test dataset (TD) and it is only used to test the networks of a given problem domain. None of the models have access to the data of this dataset.

*2) Network Trained With Original Domain (OD):* This network plays the role of the target network being copied. Since in real scenario this network would only provide a black-box API access, the original dataset used for training is confidential and not known by copycat networks. In general, the creation of this dataset is expensive for companies and access is not granted to the outside world, which increases the value of the network model. Ideally, the copycat should be able to copy this network keeping 100% of accuracy. In our experiments, one dataset is defined as the original domain dataset (ODD) and is used to train the target network. No other model have access to the data of this dataset.

*3) Network Trained with Problem Domain Data and Original Labels (PD-OL):* This network plays the role of a cheaper model created to try to replicate the original model. Although it is not easy to get access to the amount of data used to train the original model, it might be possible to get access to a smaller amount of data from the problem domain in order to train a cheaper model. This baseline model tries to replicate this situation and enables a comparison verify whether the copycat can replace the need for these cheaper models, in case they are not able to fully copy the original model. Images used to train this network are different from the ones used to train the original network, but are in the same problem domain. Note that data augmentation can be performed to increase the size of these datasets in order to enable a fair comparison, since deep models require a lot of data to be trained. In our experiments, one dataset is defined as the problem domain dataset (PDD) and is used to train the PD-OL baseline model.

*B. Copycat Networks*

Experiments consider three types of copycats: *i)* network trained with the fake dataset created with random natural images not related to the problem domain, i.e., non-problem domain with stolen labels (NPD-SL); *ii)* network trained with the fake dataset created with images from the same problem domain as the target network, i.e., network trained with problem domain and stolen labels (PD-SL); and *iii)* network trained with the fake dataset (created with images from the first) and fine-tuned with images from the second.

*1) Network Trained with Non-Problem Domain and Stolen Labels (NPD-SL):* This copycat network intends to evaluate whether the proposed method is able to copy a target model using only random natural images. In this case, it is assumed that the attacker might not have data from the same domain. Nonetheless, he still can use publicly available images from the Internet to query the target network and build a fake dataset. Therefore, samples of this dataset comprises random natural images labeled by the target network. Since these images are different from those in problem domains evaluated in this work, only one dataset was built with non-problem domain images (NPDD) and used in all experiments with NPD-SL. In total, the NPDD comprises 3,297,259 images from ImageNet (ILSVRC2017 and Visual Domain Decathlon) [15], and 123,403 images from Microsoft COCO [16]. Note that fake dataset labels (i.e., stolen labels) are generated using the target network. Therefore, these labels can be different from originals, which implicates a different distribution of images per class for each evaluated problem, as discussed in the Section V-E.

*2) Network Trained with Problem Domain and Stolen Labels (PD-SL):* This copycat network intends to evaluate whether the proposed method is able to copy a target model using a small set of images of the same problem domain as the target network. In this case, it is assumed that the attacker might have access to a small set of data from the same domain, but does not have their labels. Therefore, the samples of this dataset comprises images of the same problem domain as the original dataset, but labeled by the target network. In our experiments, the dataset used for training the PD-SL network is the same as the one used in the PD-OL network, i.e., problem domain dataset (PDD). However, it uses labels predicted by the target network instead of the original labels provided with the dataset.

*3) Network Trained with NPD-SL and fine-tuned with PD-SL (NPD+PD-SL):* This copycat network intends to evaluate whether the performance of the proposed method is improved when combining the two previous models NPD-SL and PD-SL. In this case, it is assumed that the attacker has access to publicly available images not related to the problem domain and to a small set of data from the same domain, but does not have their labels. Therefore, the NPD-SL network is further fine-tuned with the data used to train the PD-SL network.

*C. Problems*

In order to evaluate the proposed method with different real-world problems, three domains were chosen: Facial Expression Recognition, General Object Classification and Satellite Crosswalk Classification. Additionally, the method was evaluated in a publicly available API.

*1) Facial Expression Recognition (FER):* In the facial expression recognition problem, there are usually six basic universal [17] expressions that are widely used in the literature: fear, sad, angry, disgust, surprise and happy. In addition, some works in literature also take the neutral expression into consideration. In this work, seven facial expressions were used (six basic plus neutral). In the literature, there are avaiable several widely known facial expression recognition datasets. Therefore, in this work, a cross-database protocol was followed, i.e., databases for different purposes (e.g., train and test) are always different. The Extended Cohn-Kanade Database (CK+) [18] was chosen to compose the test dataset (TD). For composing the original domain dataset (ODD), the following datasets were chosen: AR Face [19], Binghamton University 3D Facial Expression (BU3DFE) [20], The Japanese Female Facial Expression (JAFFE) [21], MMI [22] and Radboud Faces Database (RaFD) [23] datasets. Finally, the Karolinska Directed Emotional Faces (KDEF) [24] was chosen to compose the problem domain dataset (PDD). Data augmentation was used as described in [25] to increase the dataset size and ensure the balance between classes. Given that most of these databases are grayscale, the NPD dataset was changed to grayscale as well, in this problem only. All networks started the training from a pre-trained model originally trained for facial recognition [26]. The number of outputs of the copycat networks was set to 7.

*2) General Object Classification (GOC):* Object classification is a highly studied problem in computer vision. There are several definitions of object types, and they are chosen according to the problem. This work followed the definition of the classical CIFAR-10 [27] and STL-10 [28] datasets that are based on 10 classes. Due to a mismatch between their classes, one class from each one was removed ("frog" from CIFAR-10 and "monkey" from STL-10), resulting in a total of nine classes for this problem. Because of that, the order of the labels was changed so ensure that the same label was referring to the same class in both datasets. The CIFAR-10 is originally divided in training and test data. Therefore, the CIFAR-10 test data was chosen to compose the test dataset (TD), whereas the CIFAR-10 training data was chosen to compose the original domain dataset (ODD). Finally, the whole STL-10 wsa used to compose the problem domain dataset (PDD). All networks started the training from a pre-trained model on the ImageNet [12]. The number of outputs of the copycat network was set to 9.

*3) Satellite Crosswalk Classification (SCC):* The Crosswalk classification problem is a binary classification problem based on satellite imagery [29]. It comprises satellite imagery from 3 continents (Europe, America, and Asia) distributed in 2 classes: crosswalk and no-crosswalk. The images were automatically collected from online APIs [29]. To compose the test dataset (TD), 15% of the images from Europe were randomly selected. The remainder 85% of the images from Europe were chosen to compose the original domain dataset (ODD). Finally, images from America and Asia were used to compose the problem domain dataset (PDD). All networks started the training from a pre-trained model on the ImageNet [12]. The number of outputs of the copycat network was set to 2.

*4) Microsoft Azure Emotion API:* An attempt to copy the performance of a publicly available cloud-based API was performed. The Microsoft Azure Cognitive Services provides an Emotion API[1], which allows an user to send a picture and receive as response the locations of the faces and "emotions" recognized for each face. This API has three pricing options: free, basic and standard tiers. The free tier allows for a reasonable amount of transactions at no cost, but it seems prohibitive for the scale of our experiments due to the number of queries necessary to generate the stolen labels. The basic and standard tiers basically differ in terms of the accepted input formats: the basic requires the user to send the location of the face in the requested image to obtain the expression, while the standard firstly localizes the faces and then recognizes the expressions (at a higher cost per transaction). As the faces are well defined in our databases, we decided to use the basic tier. The cost and time to generate required fake datasets can be seen in the Table I.

TABLE I: Microsoft Azure Pricing to Generate Fake Datasets

|  | **PD** | **NPD** |
|---|---|---|
| **Images** | $65,660$ | $3,420,662$ |
| **Cost** | $6,57 | $342,07 |
| **Time** | 01h49m | 3d23h01m |

The NPD and PD datasets were the same as the ones used in the FER problem. To label the images, as the face location was required by the basic tier (and NPD images may have no faces at all), all images were given as if the faces were the whole image. In addition, the "contempt" expression used by the API was discarded, because of the very low prediction rate by the API (less than 60 out of +3.42 million images). The number of outputs of the copycat network was set to 7.

---

[1]https://azure.microsoft.com/en-us/pricing/details/cognitive-services/emotion-api/

## D. Metrics

In order to account for possible data imbalance in the test datasets, the macro-averaged accuracy was used. The macro average computes a simple average over the classes. To measure the performance of the copycat networks, their accuracies were compared with the baselines' models. In addition to the macro-averaged accuracy, it is also reported the ratio between the performance of a given network and both the target network (OD) and the smaller problem domain network (PD-OL). These metrics are called *performance over target network* and *performance over PD-OL network*.

The perfect copycat network should achieve a macro-averaged accuracy identical to the target network (i.e., it should hit and miss the same images as the target network) and achieve 100% of performance over the target network.

## E. General Setup

All models were trained using Stochastic Gradient Descent (SGD) with a Step Down policy for the learning rate. The maximum of five epochs was chosen because the models showed convergence accuracy after this number of epochs in empirical experiments. However, the training of the target networks had more than five epochs, until their losses reached a plateau. The target network was trained with the respective original domain dataset (ODD) for 30 epochs in FER, 20 epochs in GOC, and 5 epochs in SCC. In the augmentation process, the following operations were applied: add/sub intensity, contrast normalization, crop, horizontal flip, Gaussian blur, Gaussian noise, piecewise affine, rotate, scale, sharpen, shear, and translate. These operations are combined to generate 22 types of augmentation. For each original image, each augmentation type generates 3 synthetic images. The original label is maintained for each synthetic image.

All experiments were carried out using an Intel Core i5-7500 3.40GHz with 32GiB of RAM and NVIDIA GeForce GTX 1070 with 8GiB of memory. The environment of the experiments was GNU/Linux Mint 18.1, with NVIDIA CUDA Framework 9.0 and cuDNN 7.0. The training and test phases were performed using Caffe [30]. The augmentation process used the imgaug[2] python package. The scripts, the network models with weights, and a detailed description of each dataset and experiment were publicly released[3].

## V. EXPERIMENTAL RESULTS

In this work, a process to create a copycat of a given (target) network is evaluated. In this context, three different problems were chosen to perform experiments: facial expression, object, and crosswalk classification. For each experiment, copycat networks (NPD-SL, PD-SL, and NPD+PD-SL) were compared with two baselines (OD and PD-OL), aiming to assess the capability to obtain equivalent or near-equivalent network performance using random natural images and stolen labels. The results of the experiments for each problem are presented in the next subsections, followed by a discussion of them.

[2]https://github.com/aleju/imgaug
[3]https://github.com/jeiks/Stealing_DL_Models

## A. FER

TABLE II: Performance Results for the FER Experiment

| Networks | Macro Average (accuracy) | Performance over target network | Performance over PD-OL network |
|---|---|---|---|
| OD (Target) | 88.7% | – | – |
| PD-OL | 82.5% | 93.0% | – |
| NPD-SL | 83.0% | 93.7% | 100.7% |
| PD-SL | 83.4% | 94.1% | 101.2% |
| NPD+PD-SL | 87.4% | 98.6% | 106.0% |

The FER results are presented in the Table II. As can be seen, the target network accuracy was 88.7% and the PD-OL network was 82.5%, which reached 93.0% performance over target network. Although PD-OL is able to achieve a good performance in this problem, all the other copycat networks improved over its performance.

The NPD-SL network was able to achieve 93.7% of performance over target network and 100.7% of performance over PD-OL network, reaching better accuracy than the network trained with the data from the problem domain. The PD-SL network performance was slightly better (1.1%) than the PD-OL network. The best performance was achieved by the NPD+PD-SL network, where images from the problem domain improved the NPD-SL network accuracy in 4%. When compared to the original network, it is only about 1% below in absolute accuracy.

## B. GOC

TABLE III: Performance Results for the GOC Experiment

| Networks | Macro Average (accuracy) | Performance over target network | Performance over PD-OL network |
|---|---|---|---|
| OD (Target) | 95.3% | – | – |
| PD-OL | 82.1% | 86.2% | – |
| NPD-SL | 94.0% | 98.6% | 114.4% |
| PD-SL | 90.0% | 94.4% | 109.5% |
| NPD+PD-SL | 94.0% | 98.6% | 114.4% |

The GOC results are presented in the Table III. The target network accuracy was 95.3% and the PD-OL network accuracy was 82.1%, which reached 86.2% performance over target network.

The NPD-SL network presented great results, achieving 98.6% of performance over target network and exceeded in 14.4% of performance over PD-OL network. Therefore, the copycat network trained with data from a different domain reached a performance near to the target network. The PD-SL network performance over target net was 9.5% better than the PD-OL network, which indicates the importance of the stolen labels in the copy process.

The lowest accuracy was of PD-OL network and the greatest accuracy was achieved by both the NPD-SL and NPD+PD-SL network. The PD-SL dataset did not improve the NPD-SL network in this problem domain.

TABLE IV: Performance Results for the SCC Experiment

| Networks | Macro Average (accuracy) | Performance over target network | Performance over PD-OL network |
|---|---|---|---|
| OD (Target) | 95.8% | – | – |
| PD-OL | 92.8% | 96.9% | – |
| NPD-SL | 95.0% | 99.2% | 102.4% |
| PD-SL | 94.9% | 99.1% | 102.2% |
| NPD+PD-SL | 95.1% | 99.3% | 102.5% |

## C. SCC

The SCC results are presented in the Table IV. The target network accuracy was 95.8% and the PD-OL network was 92.8%, which reached 96.9% performance over target network.

In this experiment, all networks trained with stolen labels achieved a high performance over target network with less than 1% of difference. These three networks also achieved a higher performance than the PD-OL network.

The NPD-SL network achieved a better performance than PD-OL network and 99.2% of performance over target network. The PD-SL network also achieved a high performance over target network.

The worst results came from the PD-OL network and the greatest accuracy was achieved by the NPD+PD-SL network, where the images from problem domain improved the performance by only 0.1%.

## D. Microsoft Azure Emotion API

TABLE V: Performance Results for the Microsoft Azure Emotion API

| Networks | Macro Average (accuracy) | Performance over target network |
|---|---|---|
| Microsoft Azure Emotion API | 35.1% | – |
| NPD-SL | 34.1% | 97.3% |
| PD-SL | 35.4% | 100.8% |
| NPD+PD-SL | 34.8% | 99.2% |

As can be seen, the Microsoft Azure Emotion API performed poorly (35.1%) in the test dataset when compared to the results of the FER experiments. Nonetheless, the copycat networks managed to achieve a performance very close to the API. Without using any data from the problem domain (NPD), the copycat network achieved 97.3% of the performance over the target network. Using only data from the problem domain, 100.8% of performance over the API was achieved. Surprisingly, combining both was worse (−0.6%) than using only problem domain data.

## E. Discussion

A comparison between the performance of the copycat networks and baselines is presented in the Figure 3. It can be noted that networks trained with stolen labels were able to achieve high performance over target network. Moreover, the copycat networks were also better than the PD-OL network. This indicates that, to create a copycat network of a given model, it may be more important to copy target network predictions (including its mistakes) than to use a smaller version of a real dataset (where images are correctly labeled). Therefore, stolen labels seem to be appropriate for this task.

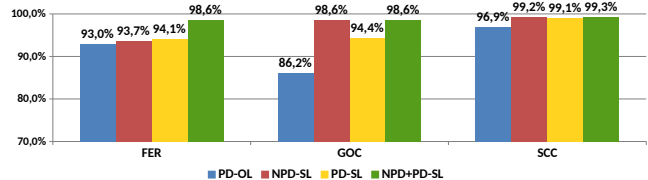

Fig. 3: Performances of all experiment networks (PD: problem domain, NPD: non-problem domain, OL: original labels, SL: stolen labels) over relative target network

NPD-SL networks show that it is feasible to create a copycat of a model of interest with high performance by just using a high number of random natural images not related to the problem domain. In addition, it showed that a large set of non-problem related images with stolen labels (NPD-SL) can be more effective in replicating the original network performance than a smaller set of problem related images with correct labels. Results with the PD-SL network showed that stolen labels can be more effective in copying the accuracy of a model than original correct labels.

The highest performance was achieved by NPD+PD-SL networks. This fact indicates that data from the problem domain helps improving the accuracy, although it may be not always worth (i.e., sometimes the cost of having an annotated labeled dataset is prohibitive and improvements are uncertain).

In the Microsoft Azure Emotion API experiment, the copycat networks managed to achieve a performance very close and better than the API, despite of the poor performance of the API on the test dataset. The results of the NDP-SL showed that random natural images are enough to copy 97.3% of the performance of the API, when evaluated on the FER test set.

In addition to the experiments described in this paper, some of the decisions were based on preliminary results. For example, the images of the NPD were converted into grayscale for the FER problem because the models trained using color images performed slightly worse ($\approx 2\%$ less of accuracy). Moreover, at an earlier attempt, we tried to ensure that the copycat networks were trained using the same distribution of samples per class as the target network. To ensure this, after being labeled by the target network, a lot of images (more than 40% from both NPD and PD) had to be discarded. Nevertheless, preliminary results showed that this was not beneficial to the copy process. Perhaps the lower number of images reduced the capacity of the copycat networks to mimic the decision frontiers of the target model. This needs further exploration.

## VI. CONCLUSION

In this paper, we presented a novel approach to create a copycat of a deep CNN model using only random unlabeled natural images. Many companies are providing cloud-based solutions powered by deep learning models. Given the necessary investments, it is their best interest to protect these models against attacks, such as the one hereby investigated.

Results showed that it is possible to achieve a performance very close to the original network (minimum of 93%, in our experiments) without using any data from the problem domain. In addition, the copycat networks were always a better alternative than training with a small database from the problem domain (i.e., they were always better than the PD-OL networks). When compared to the target network, the highest performance was achieved by NPD+PD-SL networks: 98.6%, 98.6% and 99.3% over target network for FER, GOC, and SCC. Moreover, copycat networks successfully copied at least 97.3% of the performance of the Microsoft Azure Emotion API. In conclusion, the copycat of a black-box model is feasible and companies should worry about it.

Although this investigation showed interesting results, it has some limitations. First, all the problems investigated have a low number of output classes. It would be interesting to investigate this technique with hundred or thousand classes. Second, the process of copying could be tried in a higher number of problems.

ACKNOWLEDGMENT

We gratefully acknowledge the support of NVIDIA Corporation with the donation of some of the GPUs used for this research. Cloud computing resources were provided by a Microsoft Azure for Research award.